\documentclass[preprint,12pt]{elsarticle}

\usepackage{amssymb}
\usepackage{amsmath}
\usepackage{mathtools}
\usepackage{multirow}

\usepackage{makecell}

\journal{}

\begin{document}

\begin{frontmatter}



\title{Ray-based phase error correction for miniaturized DOE projector-based FPP under single-directional hyperbolic projection}


\author[inst1]{Seung-Jae Son}
\author[inst2]{Yatong An\corref{cor}}
\author[inst1]{Jae-Sang Hyun\corref{cor}}
\cortext[cor]{Corresponding Author}

\affiliation[inst1]{organization={School of Mechanical Engineering Yonsei University},
            postcode={03722}, 
            state={Seoul},
            country={South Korea}}

\affiliation[inst2]{organization={Meta Reality Labs},
            city={Redmond},
            postcode={98052}, 
            state={Washington},
            country={USA}}

\begin{abstract}
Fringe Projection Profilometry (FPP) systems using miniaturized DOE projectors often suffer from severe phase artifacts due to nonlinear projection characteristics and limited pattern controllability. We propose a ray-based phase error correction framework that models phase artifacts along projection rays from the projector pinhole, incorporating projector geometry without relying on image-domain processing or neighboring pixels. A projector pinhole estimation method based on a single-directional hyperbolic fringe pattern is introduced, through which projector geometry can be recovered without stereo calibration. In addition, a data-efficient strategy constructs the refinement model from a single calibration pose. Experiments on miniaturized DOE projector-based FPP systems demonstrate significant improvements in reconstruction accuracy under nonlinear projection conditions, confirming the robustness and physical consistency of the proposed approach.\end{abstract}




\end{frontmatter}


\section{Introduction}

Fringe projection profilometry (FPP) is a widely used non-contact optical technique for three-dimensional (3D) shape measurement. In FPP systems, the phase encodes subpixel correspondence between the projected patterns and the captured images, enabling high-resolution surface reconstruction. Consequently, the accuracy of the reconstructed geometry is highly dependent on the precision of the recovered phase. In typical FPP pipelines, accurate phase retrieval is achieved through phase-shifting algorithms, while geometric calibration is performed to establish the mapping between phase and 3D coordinates.

Geometric calibration in FPP systems has been studied through various approaches, including stereo-based and pixel-wise formulations. In the \emph{stereo approach}, the projector is treated as an inverse camera and calibrated together with the camera using stereo calibration techniques~\cite{zhang2002flexible, zhang2006novel}. This method estimates a unified geometric model describing the relationship between image coordinates and 3D space~\cite{ma2018intrinsic, hyun2017superfast, seok2024single}. In the \emph{pixel-wise approach}, the mapping from image coordinates $(u,v)$ and phase $\Phi$ to world coordinates $(x,y,z)$ is determined independently for each camera pixel using polynomial~\cite{vargas2020hybrid,zhang2021flexible} or rational~\cite{vargas2023pixel} models, which allows finer modeling of local distortions~\cite{son2024quasi}. In addition to these approaches, alternative formulations based on illumination ray modeling have also been explored to represent the projector without relying on intrinsic parameters~\cite{yang2022novel}.

However, even after accurate calibration, reliable 3D reconstruction cannot always be achieved due to phase artifacts~\cite{pan2009phase, zuo2018phase}. These artifacts arise from non-ideal projection conditions, inter-device nonlinearities, defocus, and variations in surface reflectivity~\cite{hyun2017high, munoz2021least}. Consequently, phase error correction is essential to ensure accurate 3D reconstruction.

In most conventional FPP pipelines, system calibration and phase error correction are performed sequentially, with phase correction applied primarily in the captured image or phase map domain, without considering the projector’s physical ray structure. Typically, calibration first establishes the mapping from $(u,v,\Phi)$ to $(x,y,z)$. Subsequently, phase error correction methods—such as histogram equalization~\cite{je2021value, wang2022nonlinear}, Hilbert transform-based filtering~\cite{cai2015flexible, wang2024nonlinear, wang2025double}, or data-driven correction approaches~\cite{zhang2006generic, feng2021generalized, xu2022nonlinear}—are applied prior to reconstruction. These techniques aim to suppress periodic or localized high-frequency phase errors. However, these phase error correction methods inherently operate on the captured image or phase map and are influenced by neighboring pixel information. Moreover, these approaches often rely on assumptions that may not hold under non-ideal projection conditions. For instance, histogram equalization methods assume that phase values are uniformly distributed across the captured region, which may be violated in the presence of sharp edges, shadowed regions, or surface reflectivity variations. Similarly, Hilbert transform-based techniques assume that the phase-shifting operation used to estimate the quadrature phase is independent of surface geometry, an assumption that may not hold when strong geometric variations or optical imperfections are present.

Also, there is a growing demand for using miniaturized DOE projectors as alternatives to conventional digital projectors in fringe projection systems, particularly for applications requiring miniaturized or portable setups such as virtual reality (VR) displays~\cite{xiong2021augmented, liu2023adaptive, hirose2023chip} and mobile 3D sensing environments~\cite{li2017high, qian2019high, jeon2024motion}. Unlike digital light processing (DLP) projectors, which produce relatively stable sinusoidal fringes, miniaturized DOE projectors often generate distorted or nonlinear fringe patterns due to laser speckle, diffraction, and hardware-level modulation~\cite{kumar2021speckle, chen2025liquid}. These distortions introduce systematic phase artifacts that cannot be reliably corrected using conventional preprocessing techniques and degrade the performance of existing calibration–error correction pipelines. Consequently, there is a need for a physically grounded refinement strategy that explicitly models phase errors, correcting the systematic nonlinear distortions inherent in DOE projection systems.

Motivated by this physical insight, we propose a ray-based phase error correction framework that models phase error as a function of projection direction and radial distance from the projector pinhole. The method analyzes a 3D phase error scalar field in world coordinates and captures its structured variation along projection rays. By parameterizing points in this ray-based representation, phase errors are modeled directionally over the projector’s field of view. The resulting compensation is evaluated along each projection ray at the initially reconstructed 3D points and used to correct the corresponding measured phase.

To enable this modeling, the projector pinhole location is first estimated using a single-directional hyperbolic fringe pattern. Based on the pinhole projection assumption and geometric constraints from phase-aligned surfaces, the projector optical center is determined from the consensus intersection of multiple phase-consistent surfaces. Once the projection pinhole is estimated, phase error variation along each ray is modeled as a linear function of radial distance, with parameters defined over the angular domain.

The main contributions of this work are summarized as follows:
\begin{enumerate}
    \item A ray-based phase error correction framework that models phase artifacts along projection rays from the projector pinhole, explicitly incorporating projector geometry without relying on image-domain processing or neighboring pixel information.

    \item A novel projector pinhole estimation method using a single-directional hyperbolic fringe pattern, which allows projector geometry recovery without requiring bi-directional patterns or stereo calibration.

    \item A data-efficient correction strategy that constructs the ray-based phase error correction model from a minimal number of calibration poses, including a single-pose configuration, without additional training data or specialized procedures.

    \item Experimental validation on miniaturized DOE projector-based FPP systems, demonstrating improved reconstruction accuracy under single-directional hyperbolic and nonlinear projection conditions.
\end{enumerate}

\section{Principles}

\subsection{Conventional FPP Reconstruction and Calibration}
\label{sec:Conventional}

In a conventional FPP system, the wrapped phase $\phi(u,v)$ at each camera pixel $(u,v)$ is typically obtained from an $N$-step phase-shifting sequence as
\begin{equation}
\phi(u,v) = -\tan^{-1} \left( \frac{\sum_{k=1}^{N} I_k(u,v) \sin(\delta_k)}{\sum_{k=1}^{N} I_k(u,v) \cos(\delta_k)} \right),
\label{eq:wrapped_phase}
\end{equation}
where $I_k(u,v)$ denotes the captured intensity of the $k$-th projected fringe pattern and $\delta_k$ is its phase shift.
The wrapped phase $\phi(u,v)$ contains 2$\pi$ discontinuities due to the nature of the inverse tangent function, which are subsequently unwrapped to obtain the continuous, or absolute, phase $\Phi(u,v)$:
\begin{equation}
\Phi(u,v) = \phi(u,v) + 2\pi K(u,v),
\label{eq:absolute_phase}
\end{equation}
where $K(u,v) \in \mathbb{Z}$ is the fringe order at pixel $(u,v)$. $\Phi(u,v)$ can be converted to the corresponding projector pixel coordinate $\mathbf{p}_\text{p}$ using the known fringe period.

Two phase unwrapping strategies are commonly used in fringe projection profilometry: temporal and spatial unwrapping. In temporal phase unwrapping, multiple fringe frequencies or additional coded patterns, for example Gray codes, are projected to determine the fringe order $K(u,v)$ at each pixel, providing accurate recovery of the absolute phase $\Phi(u,v)$. In contrast, spatial phase unwrapping estimates $K(u,v)$ by exploiting local phase continuity across neighboring pixels, allowing the unwrapped phase to be reconstructed from a single set of phase-shifted fringe patterns.

With the continuous phase $\Phi(u,v)$ obtained through either temporal or spatial unwrapping, the next step is to calibrate the camera–projector pair to establish the mapping from image coordinates to 3D world coordinates. In the \emph{stereo approach}, both camera and projector are modeled as pinhole devices, with the projector treated as an inverse camera~\cite{zhang2006novel}. Stereo calibration yields their intrinsic parameter matrices $\mathbf{K}_\text{c}$ and $\mathbf{K}_\text{p}$, and extrinsic parameters $(\mathbf{R}, \mathbf{t})$. Given the camera pixel coordinate $\mathbf{p}_\text{c} = [u,v,1]^{\mathsf{T}}$ and the corresponding projector coordinate $\mathbf{p}_\text{p}$, the 3D world coordinate $(x,y,z)$ is computed via triangulation.

Pixel-wise calibration bypasses explicit pinhole modeling of the projector, instead, directly finds the relationship between $(u,v,\Phi)$ and $(x,y,z)$:
\begin{equation}
\begin{aligned}
x(u,v) &= \sum_{k=0}^{m} a^{(x)}_{k} (u, v) [\Phi(u,v)]^k,\\
y(u,v) &= \sum_{k=0}^{m} a^{(y)}_{k} (u, v) [\Phi(u,v)]^k,\\
z(u,v) &= \sum_{k=0}^{m} a^{(z)}_{k} (u, v) [\Phi(u,v)]^k,
\label{eq:polynomial_model}
\end{aligned}
\end{equation}
or with rational functions~\cite{vargas2023pixel}. For the acquisition of the reference $(x,y,z)$, pixel-wise calibration utilizes the calibrated camera parameters to reconstruct the planar calibration board over the entire image. Specifically, a planar mapping from image pixels to calibration-plane coordinates is estimated for each pose. This mapping is applied to all pixels to obtain dense plane coordinates, which are then transformed into 3D using the estimated extrinsic parameters of the board.

\subsection{Origin of Phase Artifacts Due to Fringe Nonlinearity}

In an ideal FPP system, the projected fringe pattern is assumed to be a sinusoid of the form
\begin{equation}
I(u,v) = I_0(u,v) + I_m(u,v) \cos\!\left( \phi(u,v) + \delta \right),
\label{eq:ideal_fringe}
\end{equation}
where $I_0(u,v)$ is the background intensity, $I_m(u,v)$ is the modulation amplitude, $\phi(u,v)$ is the phase to be measured, and $\delta$ is the phase shift introduced in a phase-shifting sequence.

In practice, the projector response is nonlinear due to gamma distortion, diffraction effects, or optical defocus. As a result, the projected fringe deviates from the ideal sinusoidal waveform and can be expressed as a Fourier series expansion~\cite{guo2004gamma, hoang2010generic}:
\begin{equation}
I(u,v) = I_0(u,v) + \sum_{n=1}^{\infty} I_n(u,v) \cos\!\left( n(\phi(u,v) + \delta) \right),
\label{eq:nonideal_fringe}
\end{equation}
where $n=1$ corresponds to the fundamental frequency component and $n \geq 2$ denote higher-order harmonics introduced by nonlinearity.

These higher-order harmonic components interfere with the phase-shifting demodulation process. That is, when the $N$-step phase-shifting sequence is applied, the estimated phase $\hat{\phi}(u,v)$ becomes
\begin{equation}
\hat{\phi}(u,v) = \phi(u,v) + \Delta \phi(u,v),
\label{eq:phase_error}
\end{equation}
where $\Delta \phi(u,v)$ represents the phase artifact induced by the harmonic distortion. The artifact term is typically oscillatory, spatially varying, and strongly dependent on projector characteristics~\cite{zhang2015comparative, yu2019flexible, liu2020flexible}. Consequently, the measured phase map often contains systematic distortions that must be compensated for before reliable 3D reconstruction.

In conventional 3D reconstruction pipelines, such distortions are typically addressed by applying error correction procedures to captured images or to the measured phase map $\Phi_{\mathrm{meas}}(u,v)$, producing a corrected phase map $\Phi_{\mathrm{corr}}(u,v)$:

\begin{equation}
\Phi_{\mathrm{corr}}(u,v) = \mathcal{F}\big(\Phi_{\mathrm{meas}}(u,v)\big),
\label{eq:artifact_removal}
\end{equation}

where $\mathcal{F}(\cdot)$ may represent histogram equalization~\cite{je2021value, wang2022nonlinear}, Hilbert transform filtering~\cite{cai2015flexible, wang2024nonlinear, wang2025double}, or data-driven correction methods~\cite{zhang2006generic, feng2021generalized, xu2022nonlinear}. Histogram equalization redistributes the phase values according to the cumulative distribution function of the measured phase to ensure a more uniform phase distribution across the field of view. Hilbert transform-based correction produces a 90\textdegree phase-shifted counterpart $\Phi_{\mathcal{H}}(u,v)$ of $\Phi_{\mathrm{meas}}(u,v)$ using the Hilbert transform, and the corrected phase is obtained as the mean of the original and phase-shifted components:
\begin{equation}
\begin{aligned}
\Phi_{\mathrm{corr}}(u,v) &= \frac{1}{2}\left[\Phi_{\mathrm{meas}}(u,v) + \Phi_{\mathcal{H}}(u,v)\right], \\
\Phi_{\mathcal{H}}(u,v) &= \mathcal{H}\!\big(\Phi_{\mathrm{meas}}(u,v)\big).
\end{aligned}
\label{eq:hilbert}
\end{equation}
where $\mathcal{H}(\cdot)$ denotes the Hilbert transform operator. This suppresses asymmetric nonlinear distortions by averaging the in-phase and phase-shifted components.

Although such phase error correction methods can improve reconstruction accuracy, the error cannot be fully described by the formulation above. In particular, miniaturized DOE projectors generate fringe patterns that deviate significantly from ideal sinusoidal models, unlike DLP projectors, whose projected patterns can be well characterized. As a result, the assumptions underlying conventional correction methods are not always satisfied. Moreover, these approaches operate solely in the image domain and do not explicitly account for the projection geometry. This observation motivates the development of a geometry-aware refinement strategy that explicitly accounts for the ray-dependent structure of the phase error.

\subsection{Ray-based phase error correction model}
\label{sec:Ray-based_refinement}

\begin{figure}[!htb]
  \centering
  \includegraphics[width=0.58\linewidth]{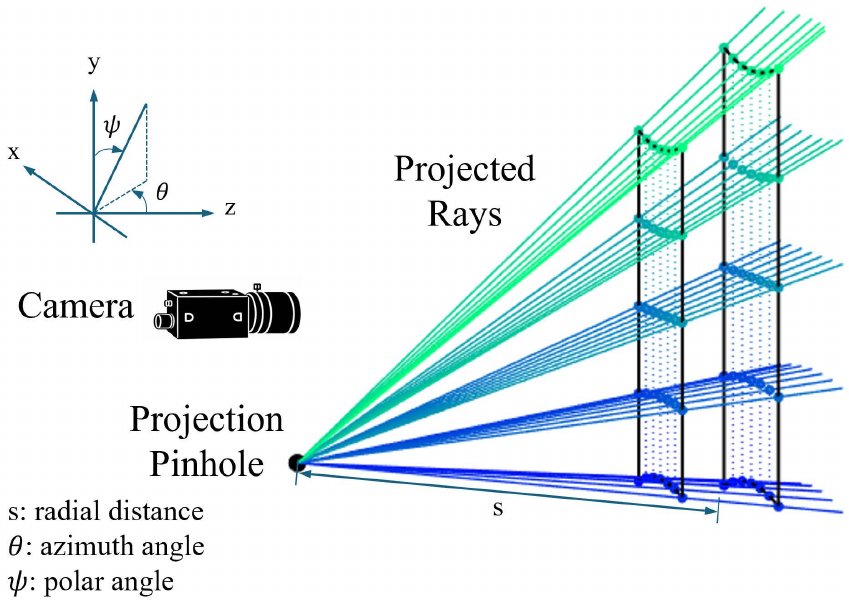}
  \caption{Schematic of the ray-based phase error correction model. The phase error is modeled along each projection ray as a function of the distance from the projection pinhole.}

  \label{fig:schematics}
\end{figure}

Phase artifacts tend to become more significant in miniaturized DOE projection systems. Once introduced, these artifacts propagate along the projection rays. Under the pinhole projection assumption—where light transport is modeled as rays emanating from a single optical center—the phase error is therefore expected to exhibit structured behavior along individual rays originating from the projector pinhole. Fig.~\ref{fig:schematics} illustrates this concept, showing rays emanating from the pinhole and the corresponding direction of phase error propagation.

To exploit this property, the phase error is analyzed with respect to projection rays, which requires accurate estimation of the projector pinhole position. In miniaturized DOE projector systems, however, the estimation of the projector pinhole is non-trivial. Unlike digital projectors that allow flexible per-pixel pattern manipulation and can determine their pinhole center via stereo-based calibration, treating the projector as an inverse camera, many DOE-based systems offer limited pattern controllability and often employ single-directional, geometrically non-linear fringe patterns. Under these conditions, methods that rely on bi-directional pattern projection, including stereo-based calibration, are not directly applicable, making pixel-wise calibration an alternative that does not provide direct access to the projector’s optical center. As a result, the pinhole location remains unknown in such models. To enable the proposed ray-based phase error correction framework, it is therefore necessary to estimate the projector pinhole position using only a single-directional hyperbolic projection pattern.

To estimate the pinhole center of the miniaturized DOE projector, three assumptions are made. First, the projector is modeled as an ideal pinhole device. Second, 3D points sharing identical phase values are assumed to lie on the same projection ray originating from the projector pinhole. Third, the projector pinhole is assumed to lie at the consensus intersection of phase-aligned hyperbolic surfaces, specifically at the vertex of the resulting parabolic locus.

\begin{figure}[!htb]
  \centering
  \includegraphics[width=\linewidth]{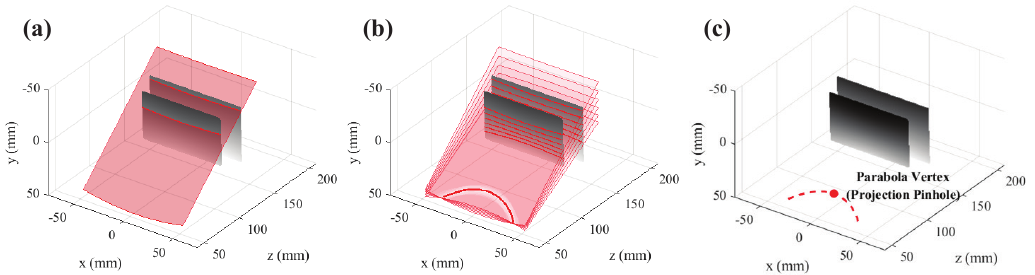}
  \caption{Schematic illustration of projector pinhole estimation with hyperbolic pattern. (a) Fitted surfaces formed by connecting $(x,y,z)$ points that share identical phase values. (b) Intersection (consensus locus) of multiple phase-aligned surfaces constraining the projector pinhole location. (c) Determination of the projector pinhole at the vertex of the parabolic consensus locus.}
  \label{fig:finding_pinhole}
\end{figure}

Under the first and second assumptions, artificial 3D surfaces are constructed by connecting $(x,y,z)$ points on the calibration board that share identical phase values in the $(x,y,z)$ coordinate system, as illustrated in Fig.~\ref{fig:finding_pinhole}(a). Since the miniaturized DOE projector pattern is hyperbolic, these phase-aligned surfaces are approximated using a second-order polynomial in the $x$-direction and a first-order polynomial in the $y$-direction. 

Because points with identical projected phase are illuminated by rays originating from a common pinhole, the pinhole must lie along each of the constructed surfaces. Repeating this process for multiple phase values yields a family of phase-aligned surfaces. The intersection, or consensus locus, of these surfaces constrains the possible location of the projector pinhole, as shown in Fig.~\ref{fig:finding_pinhole}(b). Finally, under the third assumption that the pinhole coincides with the vertex of this consensus locus, the projector pinhole position can be uniquely determined using a single-directional hyperbolic fringe pattern, as illustrated in Fig.~\ref{fig:finding_pinhole}(c).

\begin{figure}[!htb]
  \centering
  \includegraphics[width=\linewidth]{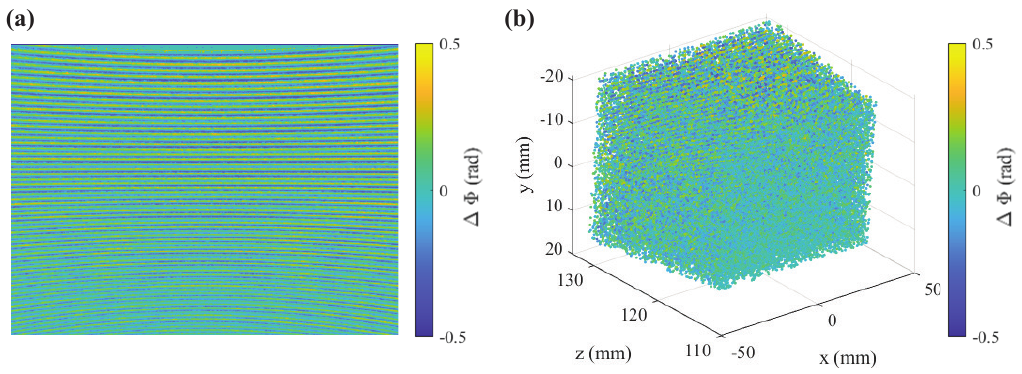}
  \caption{The phase error obtained with a calibration board. (a) Phase error obtained from a single calibration board pose. (b) 3D phase error scalar field.}
  \label{fig:phaseError}
\end{figure}

After estimating the projector pinhole position, the phase error associated with rays emanating from the pinhole was analyzed. Figure~\ref{fig:phaseError}(a) shows a sample phase error obtained from a single calibration board pose, estimated using a planar calibration board. Specifically, the calibration board used during system calibration was reconstructed into 3D geometry using the pixel-wise calibration coefficients. Since the ideal geometry of the calibration board is already available from the pixel-wise calibration process, the phase corresponding to the ideal 3D coordinate at each pixel is obtained by numerically inverting the polynomial phase-to-coordinate mapping as follows,

\begin{equation}
\Phi_{\mathrm{ideal}}(u,v)
=
\arg\min_{\Phi}
\left|
\sum_{k=0}^{m} a^{(z)}_{k}(u,v)\,\Phi^{k}
- z_{\mathrm{ideal}}(u,v)
\right|.
\label{eq:phase_inversion}
\end{equation}

The phase error is then computed as the difference between the measured phase and the ideal phase, given by
\begin{equation}
\Delta \Phi(u,v) = \Phi_{\mathrm{meas}}(u,v) - \Phi_{\mathrm{ideal}}(u,v),
\end{equation}
where $\Phi_{\mathrm{meas}}$ denotes the measured unwrapped phase and $\Phi_{\mathrm{ideal}}$ represents the phase corresponding to the ideal world coordinates of the calibration board. This phase error map serves as the fundamental input for subsequent ray-wise phase error analysis and modeling.

Using these phase error maps from all calibration poses, a 3D point cloud was obtained in world coordinates, where each reconstructed spatial point was associated with its corresponding phase error value, as illustrated in Fig.~\ref{fig:phaseError}(b).

Under the ideal pinhole projection model with optical center $\mathbf{P}_0 \in \mathbb{R}^3$, each reconstructed 3D point $\mathbf{P}_i$ lies on a projection ray emanating from $\mathbf{P}_0$. The corresponding unit projection direction is defined as
\begin{equation}
\mathbf{v}_i = \frac{\mathbf{P}_i - \mathbf{P}_0}{\|\mathbf{P}_i - \mathbf{P}_0\|},
\end{equation}
where $\|\cdot\|$ denotes the Euclidean norm. The radial distance from the pinhole is given by
\begin{equation}
s_i = \|\mathbf{P}_i - \mathbf{P}_0\|.
\end{equation}
Accordingly, each point can be expressed in ray-based form as
\begin{equation}
\mathbf{P}_i = \mathbf{P}_0 + s_i \mathbf{v}_i,
\end{equation}
which provides a physically consistent representation aligned with the projector geometry.

\begin{figure}[!htb]
  \centering
  \includegraphics[width=\linewidth]{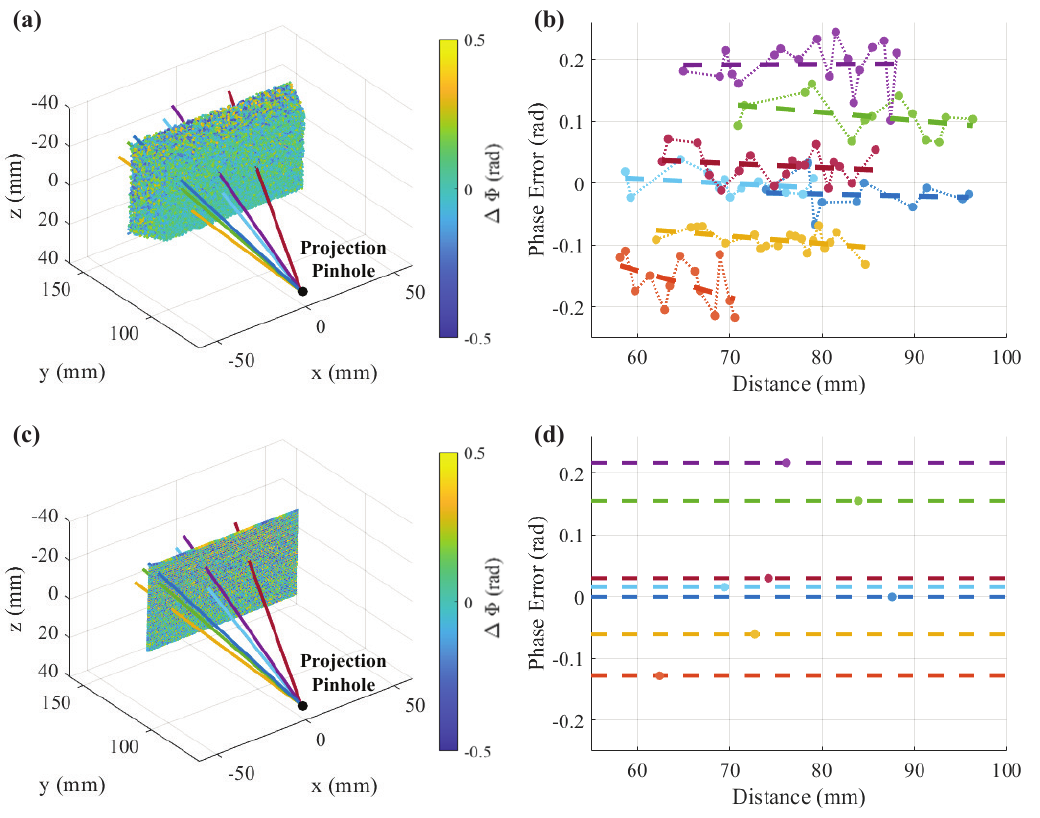}
  \caption{Ray-wise analysis of phase error with respect to the projector pinhole. 
(a) 3D visualization of the phase error field represented by colored points and selected projection rays emanating from the pinhole. 
(b) Phase error sampled along each ray as a function of distance from the pinhole, along with the fitted linear models. 
(c) Phase error distribution on a single calibration plane with the corresponding selected rays. 
(d) Constant phase error correction model for each ray, where a representative value is assigned based on the associated sampled point.}
  \label{fig:rayAnalysis}
\end{figure}

For the angular discretization of projection directions, the unit projection direction $\mathbf{v} = (v_x, v_y, v_z)^\top$ is parameterized in spherical coordinates:
\begin{equation}
\theta = \arctan2(v_y, v_x),
\end{equation}
\begin{equation}
\psi = \arccos(v_z).
\end{equation}
Here, $\theta$ and $\psi$ denote the azimuth and polar angles, respectively. This parameterization maps each ray to a point in the angular domain $(\theta,\psi)$. The angular domain is then discretized into uniformly spaced bins, allowing rays with similar directions to be grouped together. Based on this discretization, a sufficient number of representative rays were selected through angular binning of the observed directions within the projector field of view. Fig.~\ref{fig:rayAnalysis}(a) illustrates a subset of representative rays selected via this discretization process.

Under the pinhole projection assumption, systematic phase errors are expected to exhibit structured variation along each ray. Accordingly, the phase error along each ray was analyzed as a function of its radial distance from the pinhole. As shown in Fig.~\ref{fig:rayAnalysis}(b), the phase error demonstrates consistent trends along individual rays, motivating the use of ray-wise modeling. During the acquisition of phase error samples along each ray, the phase error values for each pose were interpolated to ensure continuous sampling despite the mismatch between sampling locations and measured data. A linear model was then fitted for each discretized ray to capture the variation of phase error with respect to distance, with direction-dependent coefficients $b_0(\theta,\psi)$ and $b_1(\theta,\psi)$ defined over the angular domain. To ensure robust estimation, only rays containing a sufficient number of valid observations were retained for subsequent modeling.

Once the ray-wise phase error correction model is constructed, refinement is performed directly in the reconstructed 3D space. For each reconstructed point $\mathbf{P}=(X,Y,Z)$, the direction vector relative to the projector pinhole position is computed as
\begin{equation}
\mathbf{V} = \mathbf{P} - \mathbf{P}_0,
\end{equation}

with radial distance
\begin{equation}
s = \|\mathbf{V}\|_2.
\end{equation}

The normalized direction $\hat{\mathbf{V}} = \mathbf{V}/s$ is converted into spherical coordinates $(\theta,\psi)$. Using these angular coordinates, the coefficients of the ray-based linear phase error correction model, $\{b_0(\theta,\psi), b_1(\theta,\psi)\}$, are interpolated from the coefficient maps defined over the angular domain. The predicted phase error is then evaluated as
\begin{equation}
\Delta \Phi_{\text{pred}}(\theta,\psi,s)
=
b_0(\theta,\psi) + b_1(\theta,\psi)\, s.
\end{equation}

Finally, the compensated phase is obtained by subtracting the predicted error from the measured unwrapped phase:
\begin{equation}
\Phi_{\text{comp}} = \Phi_{\text{meas}} - \Delta \Phi_{\text{pred}}.
\end{equation}

Through this procedure, phase refinement is performed along projection rays, correcting systematic nonlinear phase artifacts introduced by the projector.

\subsubsection{Iterative ray-based phase error correction}
\label{sec:iterative_refinement}

\begin{figure}[!htb]
  \centering
  \includegraphics[width=\linewidth]{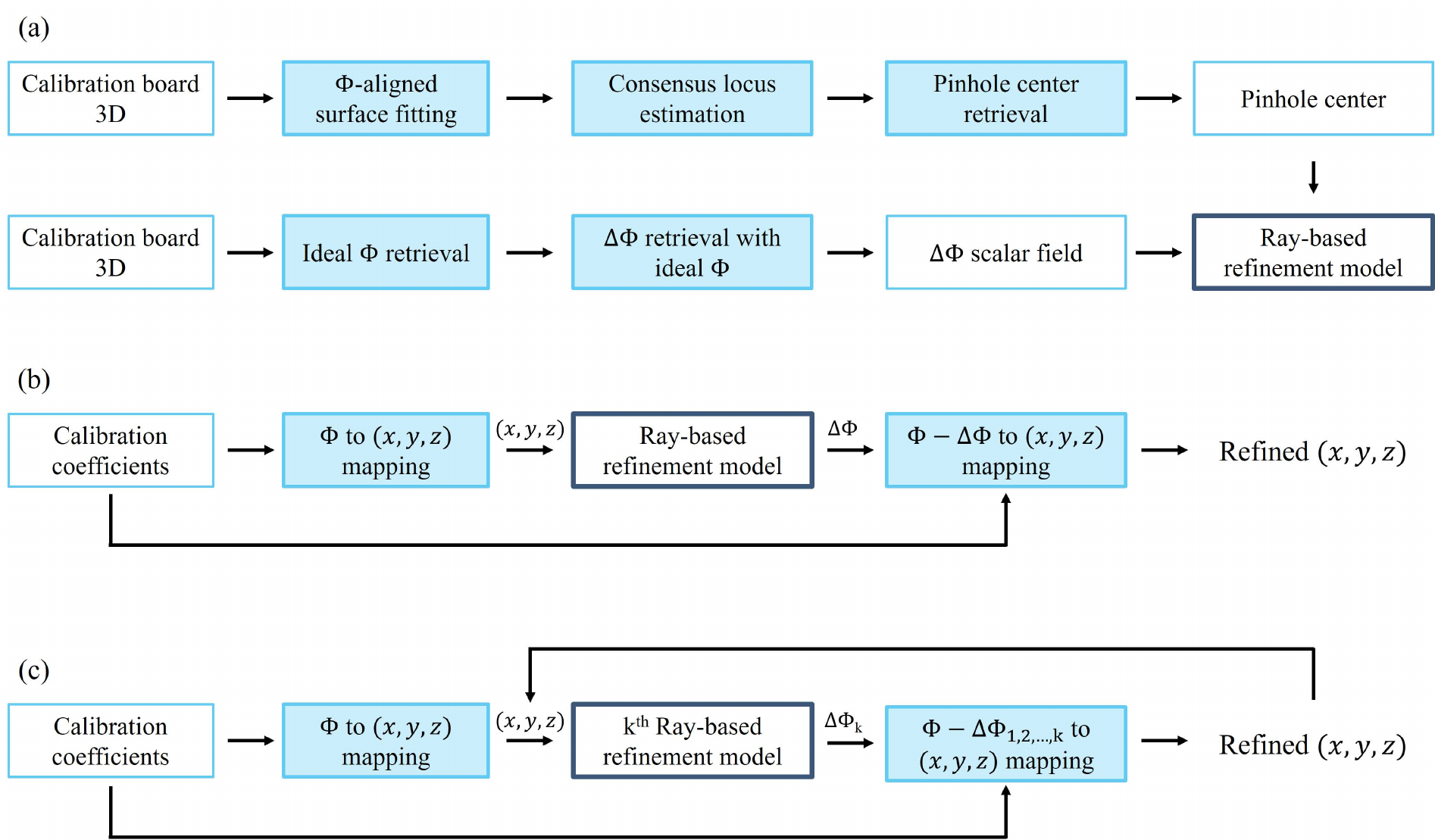}
  \caption{Workflow of the proposed phase error correction process. (a) Construction of the ray-based refinement model. (b) Standard non-iterative refinement. (c) Iterative refinement.}
  \label{fig:workflow}
\end{figure}

Although the ray-based phase error correction model described above provides a physically consistent estimate of phase artifacts, the predicted phase error depends on the initial 3D reconstructed geometry, which itself is obtained from the measured phase. Consequently, errors in the reconstructed geometry may propagate into the ray-based error prediction, although their influence is typically small. To further mitigate this coupling, an iterative refinement strategy was explored. Fig.~\ref{fig:workflow}(c) illustrates the workflow of the iterative refinement process.

In constructing the iterative refinement model, the measured unwrapped phase of the calibration boards is first converted into 3D coordinates using the pixel-wise calibration coefficients. The ray-based phase error correction is then applied using the previously constructed model, and the compensated phase is used to reconstruct an updated 3D geometry of the board.

For the subsequent iteration, the residual phase error is recomputed with respect to the known ideal calibration plane geometry. Specifically, the ideal phase is obtained by numerically inverting the polynomial phase-to-coordinate mapping, following the same procedure used in the initial modeling stage. The difference between the compensated phase from the previous iteration and the corresponding ideal phase then yields a residual phase error map. Based on this residual, a refined ray-wise phase error correction model is constructed again by sampling phase errors along projector rays and fitting a linear function of the radial distance for each projection direction.

Through this procedure, the phase error correction and ray-wise error modeling are updated in an iterative manner, progressively reducing the residual phase error while preserving consistency with the underlying projection geometry.

\subsubsection{Data-efficient refinement from a single calibration pose}
\label{sec:constant_ray_model}

Although the ray-wise formulation allows phase error to be modeled as a function of both ray direction and radial distance, empirical observations suggest that the dominant component of the systematic phase artifact is primarily determined by the projection direction itself. This observation motivates a simplified ray-consistent model in which the phase error is approximated as a constant value for each projection direction.

An important advantage of this formulation is that the ray-wise phase error correction model can be constructed using only a single calibration board pose. This enables a highly data-efficient refinement strategy that does not require multiple observations of the board. From the phase error scalar field shown in Fig.~\ref{fig:phaseError}(b), each reconstructed point $\mathbf{P}_i$ is associated with a phase error value $\Delta \Phi_i$ and a corresponding ray direction $(\theta_i,\psi_i)$. Therefore, the phase error field obtained from a single planar calibration observation already contains sufficient directional information to characterize the systematic projector-induced phase artifact.

Using representative ray directions obtained from the angular binning procedure, the phase error is modeled as,
\begin{equation} 
\Delta \Phi(\theta,\psi) \approx E_{\text{ray}}(\theta,\psi),
\end{equation}
where the phase error for each discretized direction $(\theta_k,\psi_k)$ is approximated by a single representative value $\Delta \Phi_k$ derived from observations within the corresponding angular bin.

Similar to the standard ray-based phase error correction model, a continuous representation over the angular domain is obtained by interpolating the sampled ray directions $(\theta_k,\psi_k)$ and their associated phase error values $\Delta \Phi_k$ to construct the function $E_{\text{ray}}(\theta,\psi)$. This representation enables phase error estimation for arbitrary projection directions within the projector field of view. During phase refinement, the projection direction corresponding to each reconstructed point is computed relative to the projector pinhole position. The phase error is then evaluated as $E_{\text{ray}}(\theta,\psi)$ and subtracted from the measured unwrapped phase. Through this procedure, phase artifacts are compensated consistently along projection rays while maintaining a data-efficient refinement framework.

\section{Experiments}
\label{sec:experiments}
This section presents a series of experiments conducted to demonstrate the accuracy and effectiveness of the proposed method. After introducing the experimental setup, 3D measurements on various test objects are presented for quantitative and qualitative evaluation.

\subsection{Experimental Setup}

\begin{figure}[!htb]
  \centering
  \includegraphics[width=\linewidth]{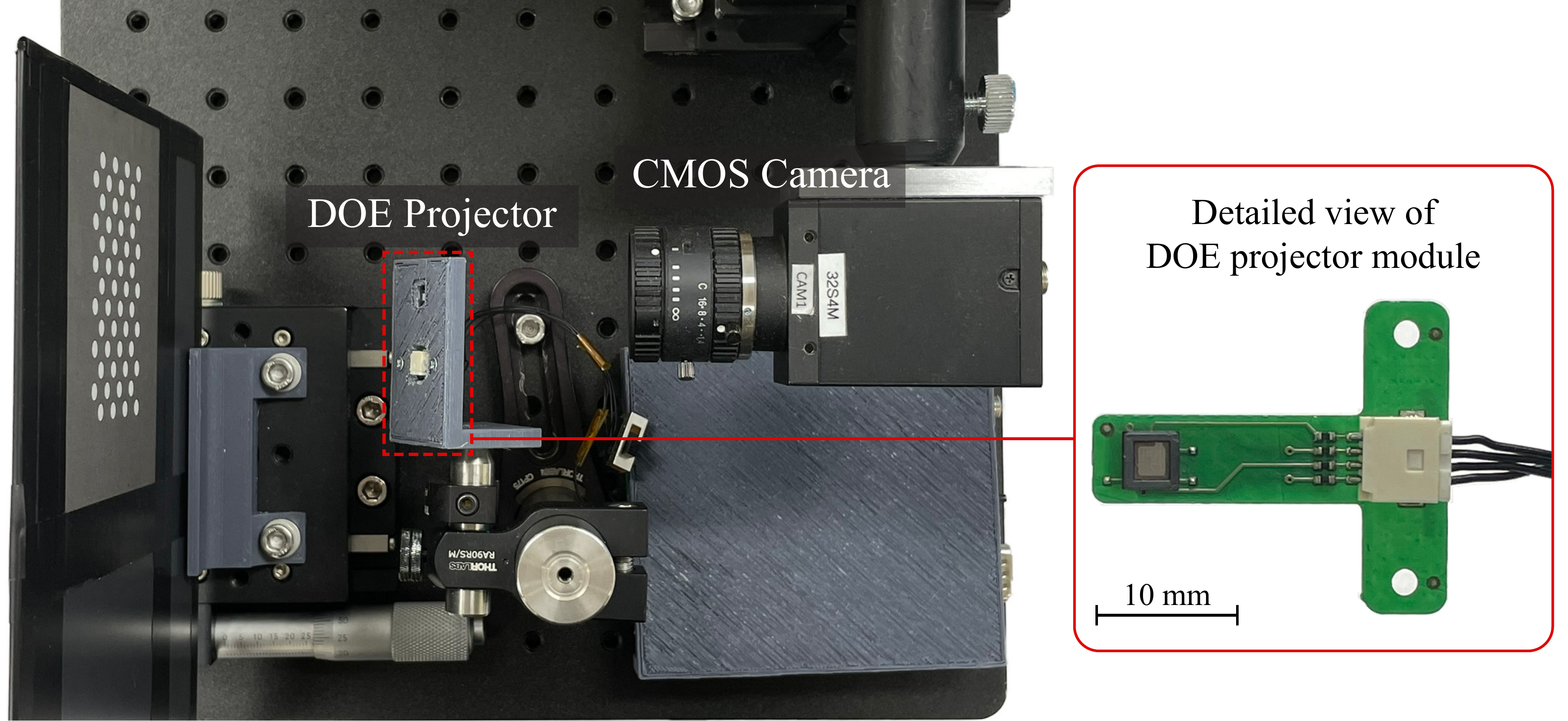}
  \caption{Experimental setup with the miniaturized DOE projector and the CMOS camera.}
  \label{fig:systemSetup}
\end{figure}

The experiments were conducted using a miniaturized DOE projector and a CMOS camera (Model: FLIR GS3-U3-32S4M-C) equipped with an 8~mm focal length lens (Model: Computar M0814-MP2). The camera resolution was set to 1440~$\times$~1080. Unlike standard DLP projectors, miniaturized DOE projectors often provide limited pattern controllability, allowing only a single-directional hyperbolic pattern to be projected, while also exhibiting nonlinear projection characteristics. Under these conditions, stereo-based projector modeling becomes impractical, and pixel-wise calibration is therefore adopted. To accurately capture these nonlinear projection characteristics, a fifth-order polynomial model was used for pixel-wise fitting, as defined in Eq.~\ref{eq:polynomial_model}. The experimental setup is shown in Fig.~\ref{fig:systemSetup}.

\begin{figure}[!htb]
  \centering
  \includegraphics[width=\linewidth]{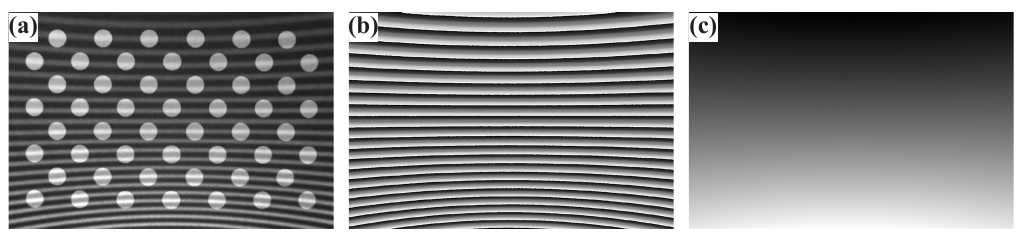}
  \caption{Sample calibration image from the miniaturized DOE projector setup. (a) Fringe pattern projected on the calibration board. (b) Wrapped phase map. (c) Unwrapped phase map.}
  \label{fig:calibImages}
\end{figure}

For system calibration, the camera was first calibrated using multiple poses of a calibration board to estimate its intrinsic parameters. A three-step phase-shifted pattern was then projected onto the calibration board using the miniaturized DOE projector for phase retrieval. The resulting wrapped phase was subsequently unwrapped using a spatial phase unwrapping method, which was consistently applied to both the calibration board and the target objects during 3D reconstruction. Fig.~\ref{fig:calibImages} shows a representative calibration pose, including the projected fringe pattern and the corresponding wrapped and unwrapped phase maps. A total of 38 calibration poses were used for pixel-wise calibration. For the construction of the standard ray-based phase error correction model, all 38 poses were used to build the 3D phase error scalar field, whereas the data-efficient variant utilized only a single representative calibration pose. During model construction, the DOE projector pinhole was estimated using the proposed pinhole estimation strategy with a single-directional hyperbolic pattern, and approximately $10^{6}$ rays were sampled through angular binning to ensure dense and uniform directional coverage. For the standard ray-based phase error model, a linear function describing the phase error with respect to radial distance was estimated for each discretized ray only when a sufficient number of valid observations were available. Specifically, rays containing fewer than five samples were excluded to ensure robust parameter estimation. In contrast, the data-efficient variant utilizes a single observation per ray and therefore does not impose this requirement. During the experiments, both the calibration board and the target objects for 3D reconstruction were positioned at distances ranging from 100~mm to 150~mm from the camera.

\subsubsection{Plane Fitting Results}

To quantitatively evaluate the improvement in reconstruction accuracy after phase error correction, a planar reference target was reconstructed in 3D. The reconstruction error was defined as the point-to-plane distance between the reconstructed points and the fitted reference plane, which was estimated using a least-squares fitting approach. The results were compared across the following methods:

\begin{itemize}
    \item Original pixel-wise calibration without phase error correction,
    \item Pixel-wise calibration with phase error correction using histogram equalization,
    \item Pixel-wise calibration with phase error correction using the Hilbert transform.
    \item Proposed data-efficient ray-based phase error correction using a single calibration pose,
    \item Proposed ray-based phase error correction with a single iteration,
    \item Proposed ray-based phase error correction with two iterations,
\end{itemize}

\begin{figure}[!htb]
    \centering
    \includegraphics[width=\linewidth]{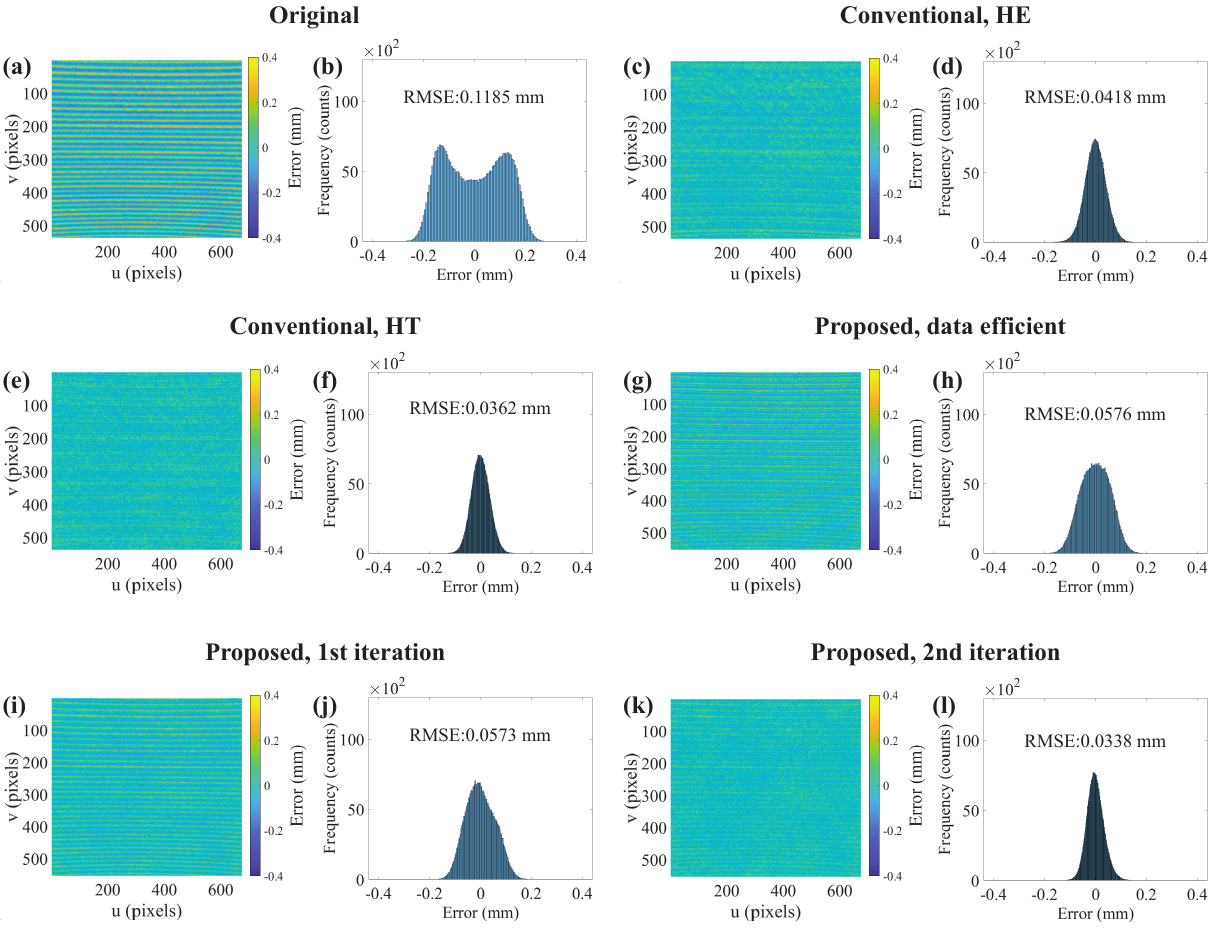}
    \caption{Plane reconstruction results. For each method, the reconstructed plane error map and the corresponding error histogram are shown. (a–b) Without phase error correction. (c–d) Histogram equalization–based phase error correction. (e–f) Hilbert transform–based phase error correction. (g–h) Proposed ray-based phase error correction using the data-efficient (single-pose) model. (i–j) Proposed ray-based phase error correction with a single iteration. (k–l) Proposed ray-based phase error correction with two iterations.}
    \label{fig:plane}
\end{figure}

The root-mean-square error (RMSE) of the plane fitting was used as the evaluation metric. Fig.~\ref{fig:plane} presents the plane fitting results along with the corresponding error maps. The original pixel-wise calibration without phase error correction yielded an RMSE of 0.1185~mm. When phase error correction was applied using conventional approaches, the RMSE was reduced to 0.0418~mm with histogram equalization and to 0.0362~mm with the Hilbert transform. The proposed data-efficient ray-based phase error correction method, which utilizes only a single calibration pose, achieved an RMSE of 0.0576~mm. When the standard ray-based model using all calibration poses was employed, the RMSE was reduced to 0.0573~mm after a single iteration, and further improved to 0.0338~mm after two iterations.

\subsubsection{Sphere Fitting Results}
\begin{figure}[!htb]
    \centering
    \includegraphics[width=0.93\linewidth]{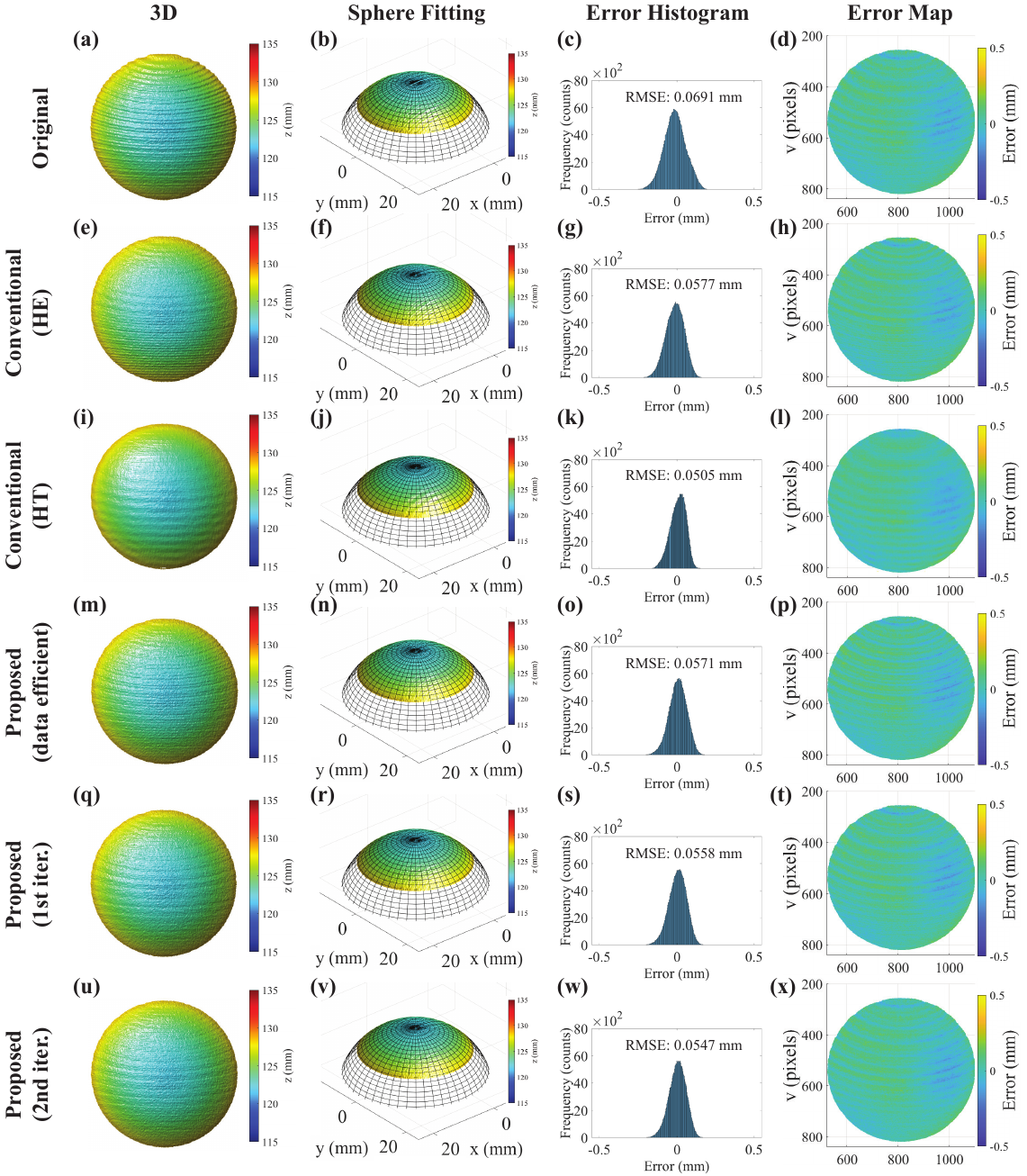}
    \caption{Sphere reconstruction results, including 3D geometry, sphere fitted geometry, error histogram, and error map. (a–d) Without phase error correction. (e–h) Histogram equalization–based phase error correction. (i–l) Hilbert transform–based phase error correction. (m–p) Proposed ray-based phase error correction using the data-efficient (single-pose) model. (q–t) Proposed ray-based phase error correction with a single iteration. (u–x) Proposed ray-based phase error correction with two iterations.}
    \label{fig:sphere}
\end{figure}

Sphere-fitting experiments were additionally conducted under the miniaturized DOE projection setup using a reference sphere with a radius of 20~mm. The reconstructed 3D surface was fitted to an ideal spherical model, and the reconstruction accuracy was evaluated by computing the root-mean-square error (RMSE) of the point-to-sphere distances. The comparison was performed using the same six methods considered in the plane-fitting experiments. Fig.~\ref{fig:sphere} illustrates the reconstructed sphere geometries along with their corresponding error distributions.

The original pixel-wise calibration without phase error correction achieved an RMSE of 0.0691~mm, with an estimated radius of 20.03~mm. When phase error correction was applied using conventional approaches, the RMSE was reduced to 0.0577~mm with histogram equalization and to 0.0505~mm with the Hilbert transform, with estimated radii of 20.03~mm and 20.04~mm, respectively. The proposed data-efficient ray-based phase error correction method, which utilizes only a single calibration pose, achieved a comparable RMSE of 0.0571~mm with an estimated radius of 20.02~mm. When the standard ray-based model using all calibration poses was employed, the RMSE was reduced to 0.0558~mm after a single iteration, with an estimated radius of 20.02~mm, and further improved to 0.0547~mm after two iterations while maintaining the same estimated radius.

Consistent with the plane-fitting results, the non-ideal projection patterns introduced strong harmonic artifacts that degraded the accuracy of the conventional pixel-wise calibration. Incorporating artifact-removal preprocessing methods, such as histogram equalization or the Hilbert transform, effectively mitigated these errors, with the Hilbert transform achieving the lowest RMSE. The proposed ray-based model achieves a comparable level of accuracy while providing a physically interpretable framework for phase error correction based on projection-ray modeling, without relying on image-domain preprocessing.

However, it should be noted that both the plane and sphere represent relatively smooth geometries, where phase variations are spatially well-behaved. Under such conditions, conventional image-domain preprocessing phase error correction methods that rely on neighboring phase relationships can perform favorably. In contrast, real-world objects often exhibit localized specularities and geometric discontinuities, which introduce strong local intensity variations and phase inconsistencies that can affect the performance of such approaches. To further evaluate the proposed method under these conditions, additional experiments were conducted, including controlled specular perturbations on the sphere and tests on geometrically complex targets exhibiting strong geometric variations, sharp edges, and shadowed regions.

\subsubsection{Sphere Fitting Results under Specular Perturbations}

\begin{figure}[!htb]
    \centering
    \includegraphics[width=\linewidth]{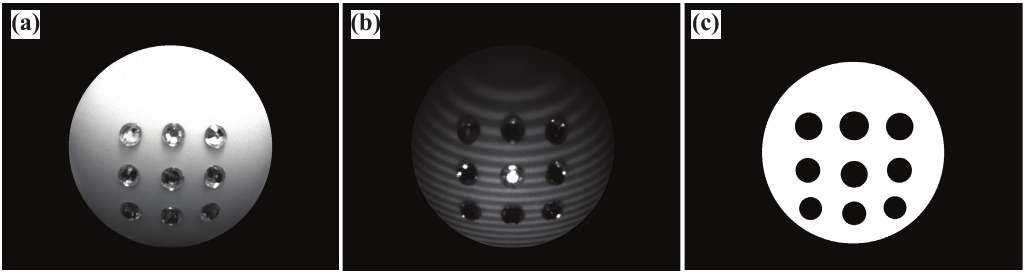}
    \caption{Experimental setup for the sphere fitting test under specular perturbations. (a) Reference sphere with attached specular beads used as the test object. (b) Captured fringe projection image. (c) Masked region (white) used for quantitative evaluation.}
    \label{fig:sphere_specular_photo}
\end{figure}

\begin{figure}[!htb]
    \centering
    \includegraphics[width=0.92\linewidth]{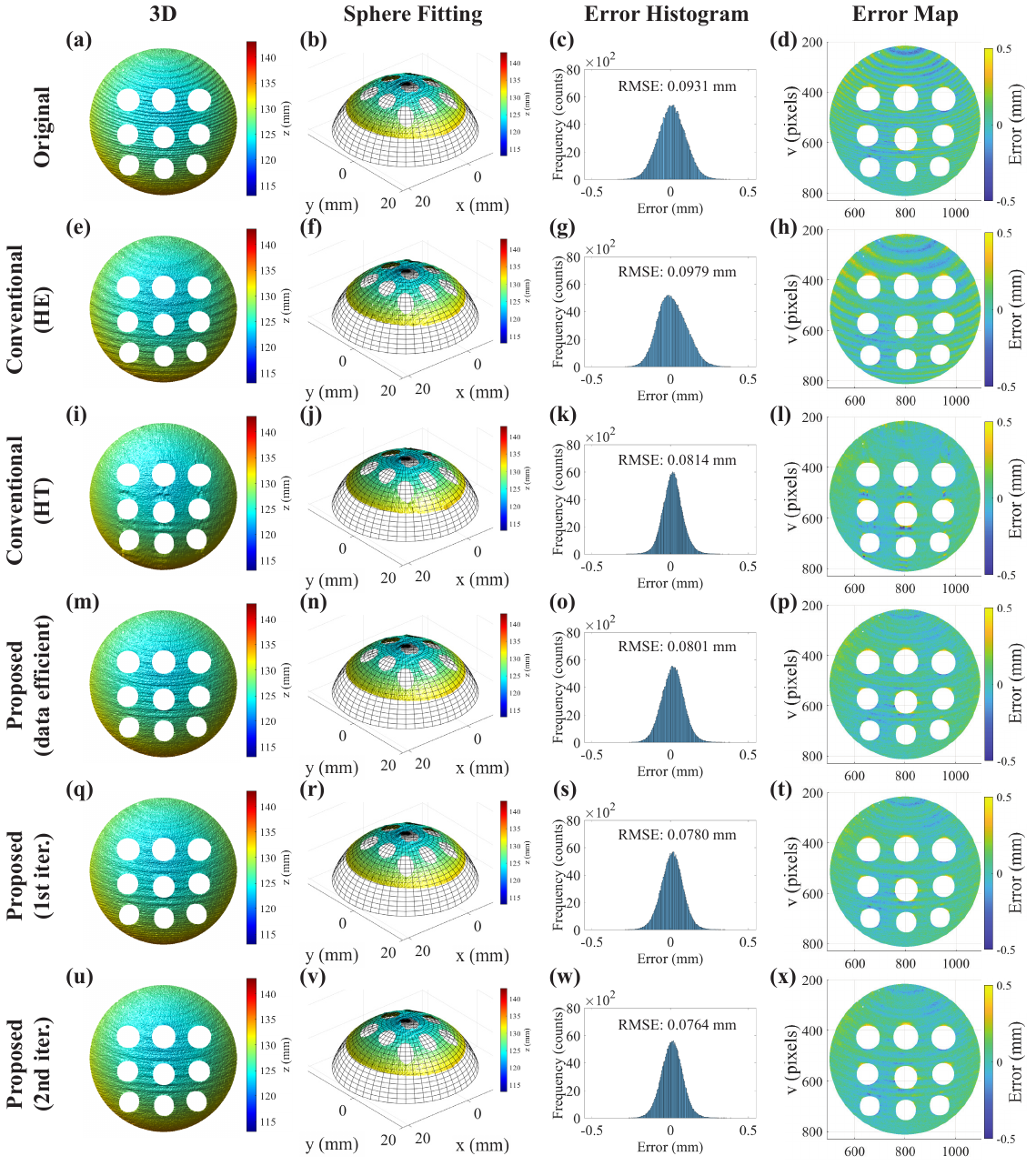}
    \caption{Sphere reconstruction results under specular perturbations, including 3D geometry, sphere fitted geometry, error histogram, and error map. (a–d) Without phase error correction. (e–h) Histogram equalization–based phase error correction. (i–l) Hilbert transform–based phase error correction. (m–p) Proposed ray-based phase error correction using the data-efficient (single-pose) model. (q–t) Proposed ray-based phase error correction with a single iteration. (u–x) Proposed ray-based phase error correction with two iterations.}
    \label{fig:sphere_specular}
\end{figure}

To evaluate performance under challenging conditions, additional sphere-fitting experiments were conducted by introducing controlled specular perturbations using reflective beads attached to the sphere surface. Fig.~\ref{fig:sphere_specular_photo} shows the test object used in the experiment. The specular perturbations introduce strong local intensity variations and phase inconsistencies, which can affect the performance of conventional image-domain preprocessing-based methods. The same evaluation protocol as in the previous sphere-fitting experiment was used.

Fig.~\ref{fig:sphere_specular} presents the reconstructed geometries and corresponding error distributions for all compared methods. For quantitative evaluation, sphere surface regions were manually selected by masking out regions affected by specular perturbations. Without phase error correction, the pixel-wise calibration yielded an RMSE of 0.0931~mm with an estimated radius of 20.04~mm. The histogram equalization method did not lead to improvement in reconstruction accuracy under these conditions, resulting in an RMSE of 0.0979~mm and an estimated radius of 20.06~mm. The Hilbert transform method demonstrated improved accuracy, reducing the RMSE to 0.0814~mm with an estimated radius of 20.06~mm.

The proposed ray-based phase error correction methods exhibited consistently improved performance under specular perturbations. The data-efficient (single-pose) model achieved an RMSE of 0.0801~mm with an estimated radius of 20.06~mm. When the standard ray-based model was applied, the RMSE was further reduced to 0.0780~mm after a single iteration, with an estimated radius of 20.05~mm, and to 0.0764~mm after two iterations, with an estimated radius of 20.04~mm.

Unlike the smooth sphere case, the presence of specular perturbations introduces localized phase inconsistencies that violate the assumptions of conventional image-domain correction methods. As a result, the performance of the conventional image-domain phase error correction methods becomes less consistent. In contrast, the proposed ray-based approach maintains robustness by modeling phase errors along projection rays, reducing the influence of local intensity anomalies and improving reconstruction accuracy.

\subsubsection{PCB Plane Fitting Results}

\begin{figure}[!htb]
    \centering
    \includegraphics[width=\linewidth]{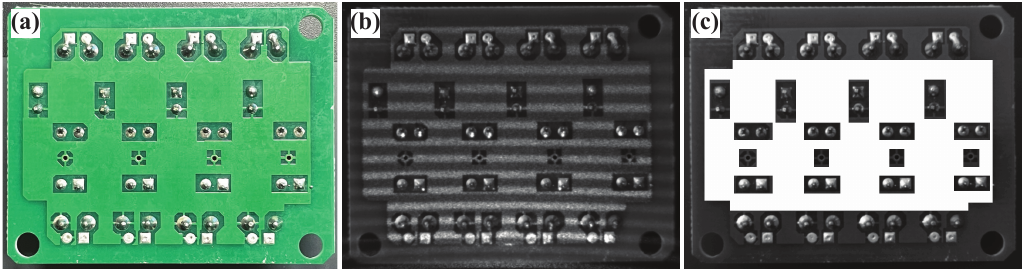}
    \caption{Experimental setup for the PCB accuracy test. (a) Printed circuit board (PCB) used as the test object. (b) Captured fringe projected image. (c) Masked region (white) used for quantitative evaluation.}
    \label{fig:pcb_photo}
\end{figure}

To further evaluate the phase error correction methods, additional quantitative evaluations were conducted using a printed circuit board (PCB) plate. The PCB surface contains multiple planar regions, along with sharp discontinuities and specular reflections, posing significant challenges for conventional image-domain preprocessing methods. Fig.~\ref{fig:pcb_photo} shows a photograph of the PCB used in the experiment.

\begin{figure}[!htb]
    \centering
    \includegraphics[width=0.97\linewidth]{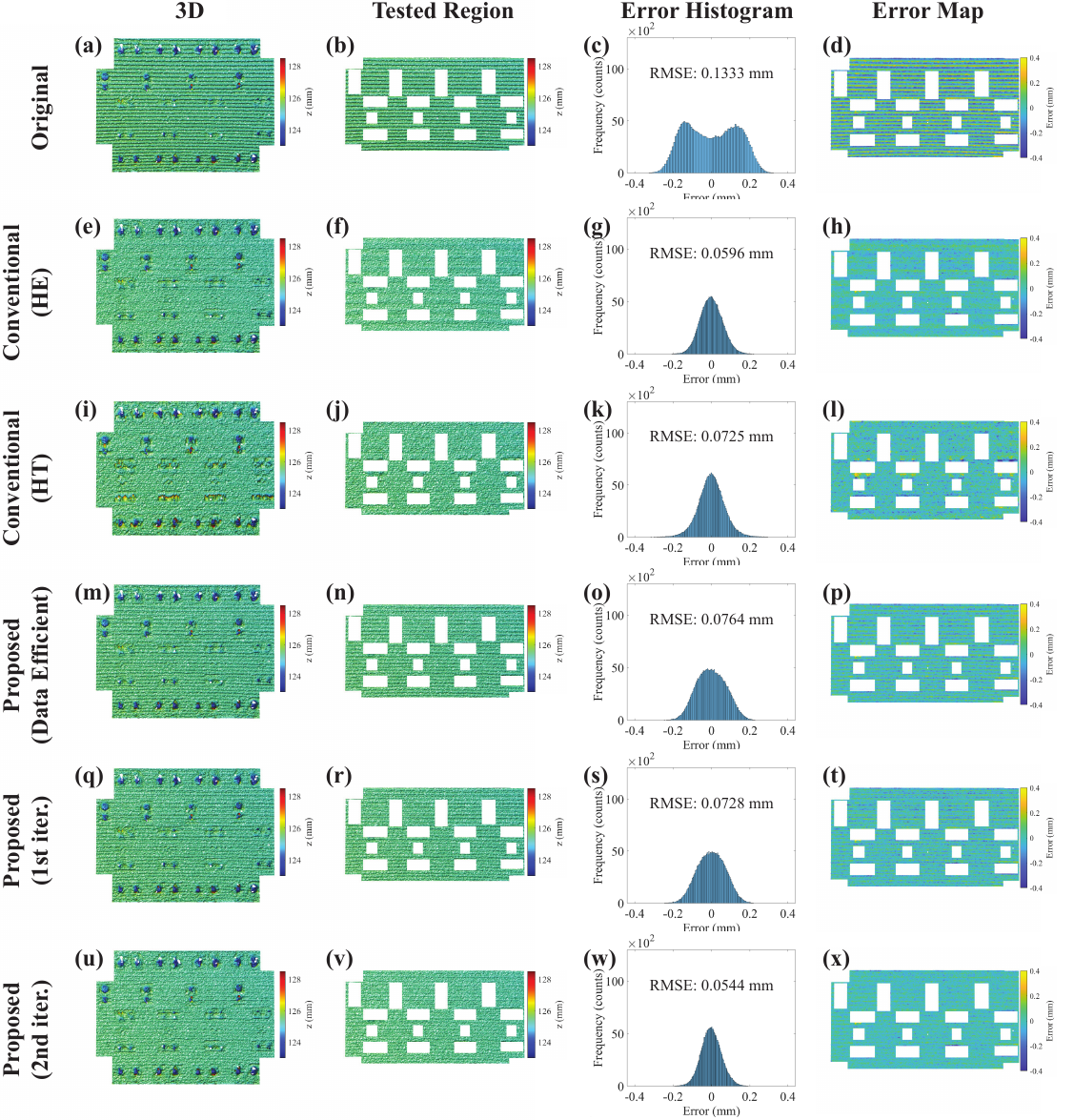}
    \caption{3D reconstruction of PCB, including 3D geometry, geometry of the tested region, error histogram, and error map. (a–d) Without phase error correction. (e–h) Histogram equalization–based phase error correction. (i–l) Hilbert transform–based phase error correction. (m–p) Proposed ray-based phase error correction using the data-efficient (single-pose) model. (q–t) Proposed ray-based phase error correction with a single iteration. (u–x) Proposed ray-based phase error correction with two iterations.}
    \label{fig:pcb_Laser}
\end{figure}

For quantitative evaluation, planar regions on the reconstructed PCB surface were manually selected, and plane fitting was performed. The reconstruction accuracy was assessed by computing the root-mean-square error (RMSE) of the point-to-plane distances. The same six methods used in the previous experiments were considered for comparison. Fig.~\ref{fig:pcb_Laser} illustrates the reconstructed PCB geometries along with their corresponding error distributions.
The original pixel-wise calibration without phase error correction resulted in an RMSE of 0.1333~mm. When phase error correction was applied using conventional approaches, the RMSE was reduced to 0.0596~mm with histogram equalization and to 0.0725~mm with the Hilbert transform. The proposed data-efficient ray-based phase error correction method, which utilizes only a single calibration pose, achieved an RMSE of 0.0764~mm. When the standard ray-based model using all calibration poses was employed, the RMSE was reduced to 0.0728~mm after a single iteration and further improved to 0.0544~mm after two iterations.

The proposed method achieved the lowest RMSE in this experiment, particularly after iterative refinement. This result can be attributed to the fact that the PCB surface contains strong geometric variations and specular reflections, under which conventional image-domain preprocessing methods are less effective in preserving geometric consistency. In particular, regions near PCB components tend to exhibit larger reconstruction errors for the Hilbert-transform-based method, likely due to its reliance on image-domain processing, which can introduce local distortions and propagate errors across neighboring pixels. Experimental results with PCB indicate that the proposed ray-based model provides robust phase error correction across diverse surface characteristics, especially in scenarios involving complex geometry and non-uniform reflectance, where physically grounded modeling becomes critical.

\subsubsection{3D Reconstruction of a Complex Object}

\begin{figure}[!htb]
    \centering
    \includegraphics[width=\linewidth]{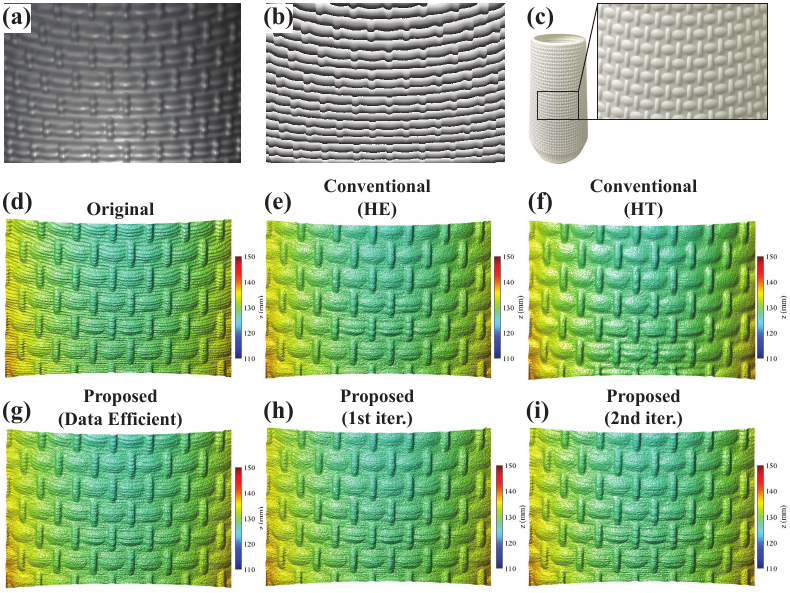}
    \caption{3D reconstruction of an embossed surface. (a) Fringe projected image. (b) Wrapped phase map. (c) Embossed surface. (d) Reconstruction without phase error correction. (e) Histogram equalization–based phase error correction. (f) Hilbert transform–based phase error correction. (g) Proposed ray-based phase error correction using the data-efficient (single-pose) model. (h) Proposed ray-based phase error correction with a single iteration. (i) Proposed ray-based phase error correction with two iterations.}
    \label{fig:texture}
\end{figure}
\begin{figure}[!htb]
    \centering
    \includegraphics[width=0.9\linewidth]{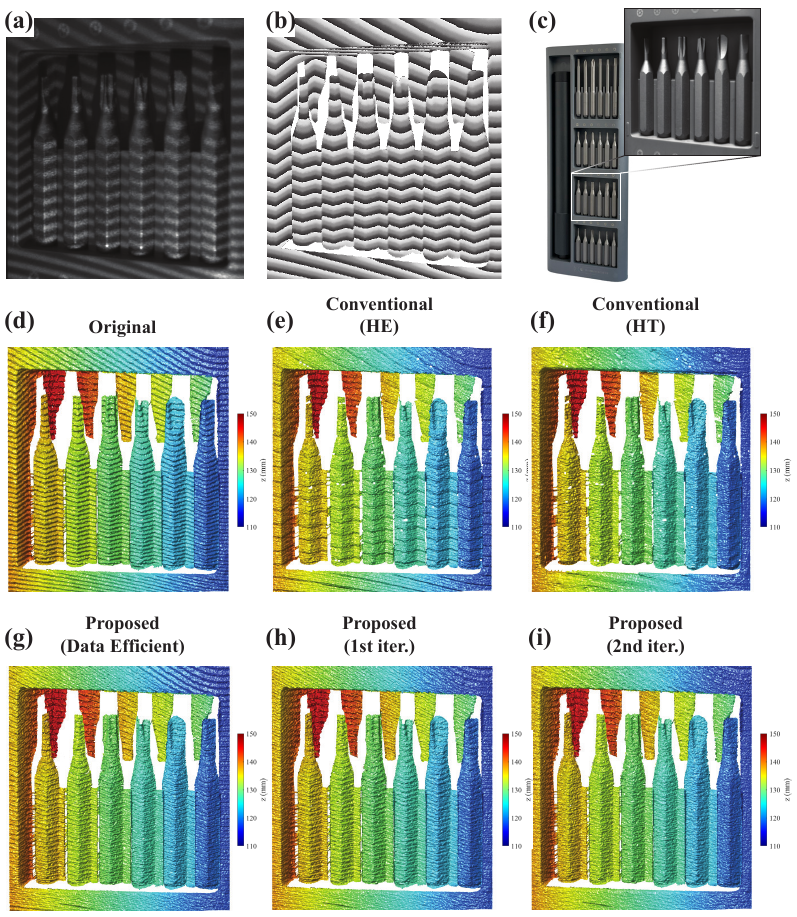}
    \caption{3D reconstruction of a screwdriver set. (a) Fringe projected image. (b) Wrapped phase map. (c) Photograph of the screwdriver set. (d) Reconstruction without phase error correction. (e) Histogram equalization–based phase error correction. (f) Hilbert transform–based phase error correction. (g) Proposed ray-based phase error correction using the data-efficient (single-pose) model. (h) Proposed ray-based phase error correction with a single iteration. (i) Proposed ray-based phase error correction with two iterations.}
    \label{fig:driver_Laser}
\end{figure}
Finally, qualitative experiments were conducted on two representative objects, an embossed surface and a screwdriver set, using the miniaturized DOE projector. Consistent with the previous observations, the non-ideal projection patterns introduce systematic phase artifacts that degrade the reconstruction accuracy of the original pixel-wise calibration. As shown in Fig.~\ref{fig:texture}, conventional methods produce visually reasonable results on the embossed surface, which exhibits relatively smooth geometry and gradual surface variations. However, their performance degrades noticeably for the screwdriver set(Fig.~\ref{fig:driver_Laser}), which contains sharp edges, fine structures, and shadowed regions. In such cases, image-domain preprocessing methods tend to be less effective at resolving phase artifacts. In contrast, the proposed method reconstructs detailed surface geometry with reduced phase artifacts, producing more geometrically consistent results across both objects. Overall, these results demonstrate that the proposed approach provides improved phase error correction performance, particularly for complex geometries and challenging surface conditions.

\section{Conclusion}
  
In this paper, a ray-based phase error correction framework was proposed for fringe projection profilometry (FPP) systems using a miniaturized DOE projector. Unlike conventional image-domain approaches, the proposed method models systematic phase artifacts along projection rays from the projector pinhole, which allows the phase error correction within a physically interpretable framework. A data-efficient refinement strategy was also introduced by simplifying the ray-wise model into a direction-dependent form that can be constructed from a single calibration pose. The framework was validated through both quantitative and qualitative experiments. For smooth reference objects such as a plane and a sphere, conventional phase error correction methods effectively mitigated phase artifacts, while the proposed method demonstrated a comparable level of accuracy. Under more challenging conditions involving specular perturbations and sharp geometrical discontinuities, the proposed ray-based approach showed improved reconstruction accuracy compared to conventional methods. Qualitative evaluations on geometrically complex objects further indicate the robustness of the proposed method, suggesting more stable and structurally consistent reconstructions. These results indicate that the framework remains effective even under the miniaturized DOE projector setup, characterized by strong spatial nonlinearity and an unmodeled projection pattern. Overall, the proposed framework provides a physically interpretable and effective solution for systematic phase artifact correction in hardware-limited projection environments. It establishes a solid foundation for future research on real-time phase refinement and 3D reconstruction, particularly for miniaturized projection systems designed for portable and mobile 3D sensing applications.

 \bibliographystyle{elsarticle-num} 
 \bibliography{cas-refs}

\end{document}